\documentclass[runningheads]{llncs}
\usepackage{graphicx}
\usepackage{subcaption}
\usepackage{amsmath}
\usepackage{mathtools}
\usepackage{multirow}
\usepackage{comment}
\begin{document}
\title{Colonoscopy Landmark Detection using Vision Transformers}
%
%
\author{Aniruddha Tamhane\inst{1} \and Tse'ela Mida \inst{1} \and Erez Posner \inst{1} \and Moshe Bouhnik \inst{1}\orcidID{0000-0003-3003-2604}}
\authorrunning{Aniruddha Tamhane et al.}
%
\institute{Intuitive Surgical, Inc. \\ 1020 Kifer Road, Sunnyvale, CA \\ \email{\{aniruddha.tamhane, tseela.mida, erez.posner, moshe.bouhnik\}@intusurg.com}}

\maketitle              
\begin{abstract}
 Colonoscopy is a routine outpatient procedure used to examine the colon and rectum for any abnormalities including polyps, diverticula and narrowing of colon structures. 
A significant amount of the clinician's time is spent in post-processing snapshots taken during the colonoscopy procedure, for maintaining medical records or further investigation. Automating this step can save time and improve the efficiency 
of the process. In our work, we have collected a dataset of 120 colonoscopy videos and 2416 snapshots taken during the procedure, that have been annotated by experts. Further, we have developed a 
novel, vision-transformer based landmark detection algorithm that identifies key anatomical landmarks (the appendiceal orifice, ileocecal valve/cecum landmark and rectum retroflexion)
from snapshots taken during colonoscopy. Our algorithm uses an adaptive gamma correction during preprocessing to maintain a consistent brightness for all images.
We then use a vision transformer as the feature extraction backbone and a fully connected network based classifier head to categorize a given frame into four classes: the three landmarks or a non-landmark frame. We compare the vision transformer (ViT-B/16) backbone with ResNet-101 and ConvNext-B backbones that have been trained similarly. 
We report an accuracy of 82\% with the vision transformer backbone on a test dataset of snapshots.

\end{abstract}
\section{Introduction}
Colorectal cancer (CRC) is among the leading causes of death worldwide \cite{chen2020cause}. In the United States alone, 161,470 individuals are estimated to be diagnosed with CRC and 54,250 individuals
are estimated to die from CRC in 2022 \cite{siegel2022cancer}. Colorectal cancer incidence rates have been increasing among screening-age individuals aged 65 years and older by 1\% per year \cite{siegel2020colorectal}.
Early onset CRC rates have also been on the rise among the patients under the recommended screening age (50 years). Early screening for colorectal abnormalities is associated with a 67\% reduction in mortality from CRC \cite{Doubeni291}. Colonoscopy being the gold standard for CRC screening \cite{issa2017colorectal} plays a critical role in mitigating risk.
\\ Snapshots taken during the colonoscopy are a critical yet time-consuming part of the post-procedural diagnosis and documentation. Physicians typically take snapshots of key colon landmarks such as the Appendiceal Orifice (AO), Ileocecal Valve (ICV),
Cecum landmark (Cec) and certain findings such as polyps, diverticula, or routine procedural steps such as a Rectum Retroflexion (RecRF), as recommended by the American Gastroenterological Institute \cite{cooper2020use}. The snapshots are useful in the post-procedural phase to serve as a medical record 
of the highlights of the colonoscopy and the patient's colonic health or for assessing the extent of the procedure by capturing a snapshot of the appendiceal orifice and ileocecal valve \cite{morelli2010colonoscopy}.
\\ It has been reported in \cite{mcdonald2014use} that a significant amount of a clinician’s time is spent maintaining Electronic Health Records. With the increase in demand for colonoscopy procedures, there is a need for improving the efficiency to save the colonoscopy clinician's time. 
There have been multiple robust, highly accurate and efficient approaches developed for polyp detection  \cite{mamonov2014automated}, \cite{park2012colon}, \cite{qadir2021toward}. However, there has been a limited amount of 
research on landmark detection. To the best of our knowledge, the algorithms developed by \cite{cao2006automatic} and \cite{lebedev2020automatic} have been the only attempts at detecting the 
appendiceal orifice (using classical and deep learning techniques respectively). The deep-learning technique developed by \cite{jheng2021novel} to detect the hepatic and splenic flexure, is the only multi-landmark detection algorithm for colons. 
We believe that this scarcity of available literature may be due to a lack of availability of expert annotated datasets of colon landmarks and the inherent difficulty of the task due to: \textit{1)} intra-colon (patient) similarity between different regions, \textit{2)} inter-colon (patient) variability in the anatomical structures of the same region of the colon and \textit{3)} non-ideal photometric conditions of the snapshots (due to poor focus, blur, reflections on the colon walls, occlusions by fluids, polyps etc.)
Thus, there is a need for developing a robust technique that can accurately identify anatomical landmarks in the colon across multiple patients, that has been rigorously tested on a dataset containing colonoscopy 
snapshots that are representative of the typical clinical setting. Further, it is important to design a data-efficient training framework that can demonstrably generalize across different anatomies.
\\ We propose a vision transformer based training framework that enables a model trained on videos (which are cheaper to annotate) to be adapted for snapshots. In our work, we address  the following problems pertaining our task: \textit{1)} adaptation to differences in data distribution from video-annotations to snapshots \textit{2)}  extreme class imbalance, \textit{3)} poor photometric conditions and \textit{4)} inconsistent annotations from experts.

\section{Related work}
A large body of work on the application of statistical, physics-based analysis and machine-learning techniques on colonoscopy has accumulated over the years primarily focusing on the detection of polyps and to a lesser extent, colon landmarks. We review the following categories of scientific literature relevant to our work:
\subsection{Landmark detection}
Fast, reliable techniques of detecting anatomical landmarks are crucial to medical image analysis. Landmark detection in ultrasound and CT scans is a well explored field, with research on detecting landmarks to utilizing them for
organ segmentation \cite{chowdhury2008detection}, \cite{ghesu2016artificial}, \cite{zhou2021review}, \cite{zhou2020landmark}. Detecting landmarks in endoscopy and colonoscopy has a smaller yet broader research focusing on identifying 
different landmarks and regions as a part of the endo-/colonoscopy process. In \cite{cao2006automatic}, a shape-based feature extraction model combined with K-Means clustering was used to detect the appendiceal orifice in colonoscopy
videos. Since this method relies on edge-based shape detection, there is a possibility of it not working on blurry images, which are characteristic of typical colonoscopy snapshots. A deep-learning based
approach was proposed in \cite{adewole2020deep} for detecting the anatomical regions (e\,.g. stomach, oesophagus etc.) from capsule endoscope frames. This demonstrated the efficacy of deep networks to correctly identify anatomical 
regions from a single endoscopy frame. The first major attempt at identifying certain colon landmarks from colonoscopy frames using deep neural networks was made by \cite{che2021deep}. They trained a large 2D CNN based neural network 
to classify a given frame as either one of splenic flexure, hepatic flexure or sigmoidal colon junction. Their approach relies on removing blurred
frames using a heuristic, and on testing the model on non-overlapping frames from the videos common to the training set. 

\subsection{Visual feature backbones and Optimizers}
Convolutional Neural Network based architectures such as the VGG-16 \cite{simonyan2014very} and ResNet101 \cite{he2016deep} have traditionally been the most effective and widely used visual feature extraction architectures. The ConvNext \cite{liu2022convnet} is the latest state-of-the-art CNN-based architecture.
On the other hand, the transformer architecture \cite{vaswani2017attention}, which is the standard architecture in Natural Language Processing, has now been adapted for vision-related  tasks in \cite{dosovitskiy2020image} showing promising results. Due to the fundamentally different mechanisms of transformer-based (attention) and  CNN-based architectures (learned filters), we decide to compare both types of architectures for our task. For our primary model, we use a Vision Transformer pre-trained on the ImageNet dataset as the visual feature extraction backbone. We also independently train a ResNet-101 and a ConvNext based model for comparison.
The choice of optimizer used directly affects the optimization landscape impacting the accuracy and ability to generalize, as show in \cite{chen2021vision}. We use a Sharpness Aware Minimization (SAM) \cite{foret2020sharpness} approach to optimizing neural networks due to its positive impact on the accuracy as well as producing semantically meaningful attention maps in case of transformers. 

\section{Data collection}
We have collected and annotated 120 colonoscopy videos and 2416 snapshots that have been used for training and evaluating our algorithm respectively. We describe the annotation process, training dataset and snapshots dataset in the following subsections.

\subsection{Annotations and cross-validation}
\label{sec:annot_cv}
We have annotated the videos on a frame-level and have cross-validated the annotations between the medical experts. This ensures a clinically accurate dataset that has fine-grain annotations with fewer human errors. We have followed the same procedure while annotating the training videos as well as the snapshots dataset. Our annotation methodology is as follows: we separate videos for the training data (which will be further split into validation and testing sets) and the snapshots dataset. Separating the data on a video-level is critical to ensure that the model generalizes well to all the anatomical variations found in colons. Each of the videos in the training datasets is then labelled on a frame-level by two medical students independently. Only the frames with a consensus between the two annotators are chosen for training and the rest are discarded.
On the other hand, each of the videos in the snapshots dataset was examined by a senior medical expert to extract snapshots, as they would in a clinical setting. Each of these snapshots was then labelled independently by two senior medical experts, and a similar consensus-based cross-validation heuristic was used to select the snapshots with matching annotations from the two experts.

\begin{table}
\centering
\caption{Snapshots and test dataset label distribution}\label{snapshot_table}
\begin{tabular}{|l|l|l|}
\hline
Label &  Number of frames (Snapshot) & Number of frames (test)  \\
\hline
Appendiceal Orifice &  518 & 776\\
Ileocecal Valve/Cecum Landmark &  132 & 133\\
Rectum retroflexion & 716 & 140\\
Other & 1050 & 1488\\
\hline
\end{tabular}
\end{table} 

\subsection{Snapshots dataset}
\begin{figure}
\centering
\begin{subfigure}[b]{0.3\textwidth}
  \centering
  \includegraphics[width=.7\linewidth]{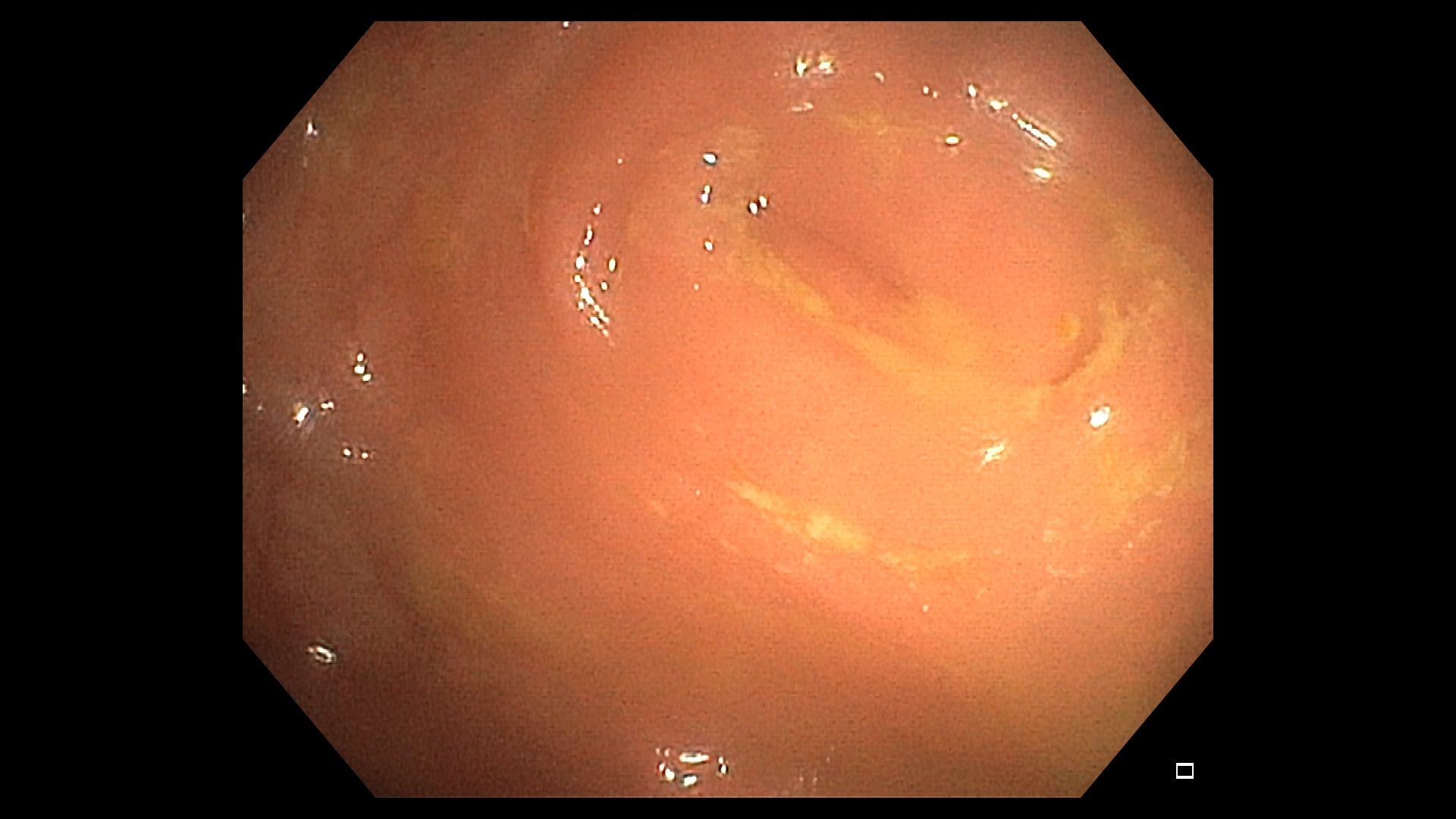}
  \caption{}
  \label{fig:sample_ao}
\end{subfigure}
\begin{subfigure}[b]{0.3\textwidth}
  \centering
  \includegraphics[width=.7\linewidth]{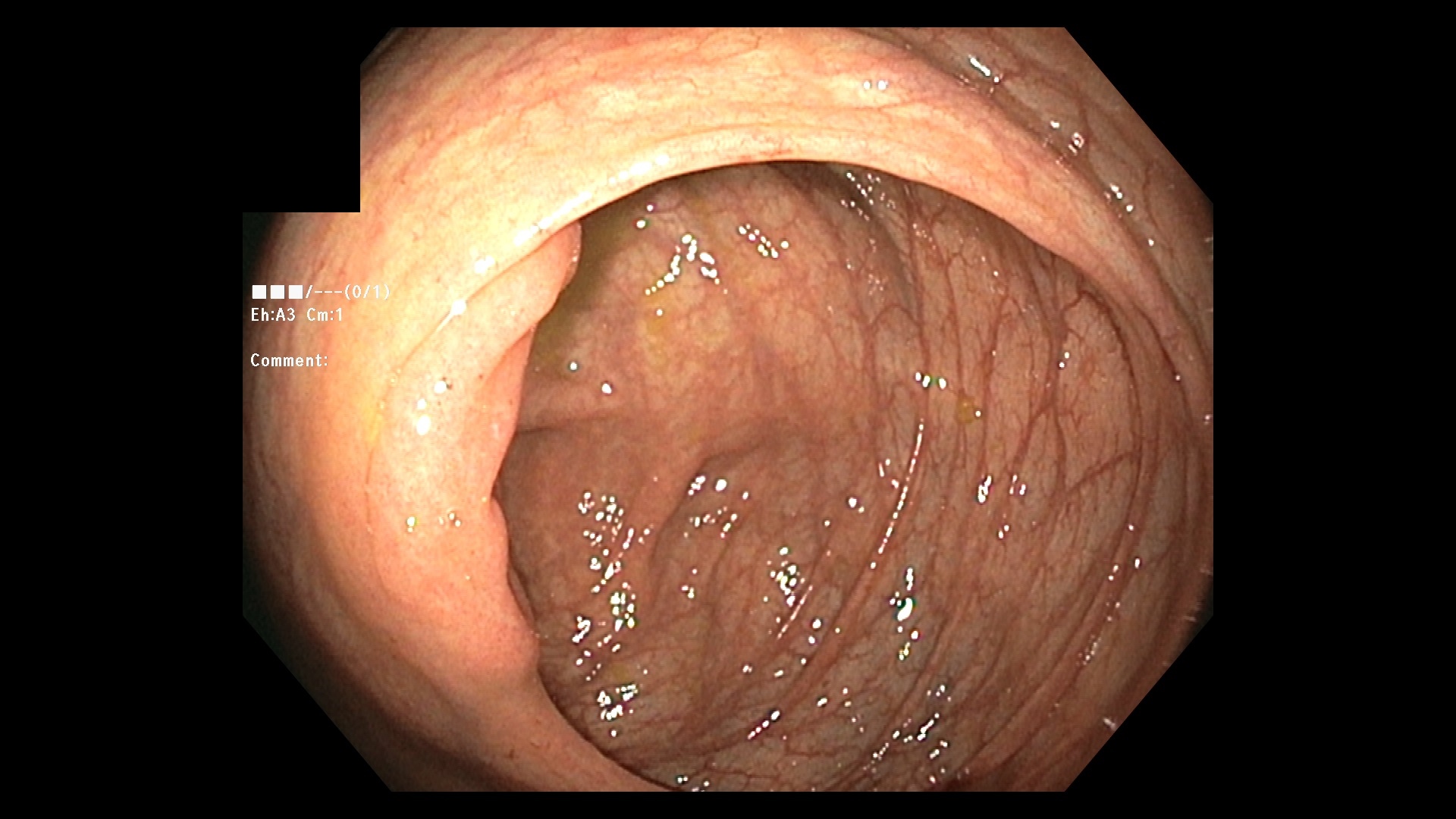}
  \caption{}
  \label{fig:sample_icv}
\end{subfigure}
\begin{subfigure}[b]{0.3\textwidth}
  \centering
  \includegraphics[width=.7\linewidth]{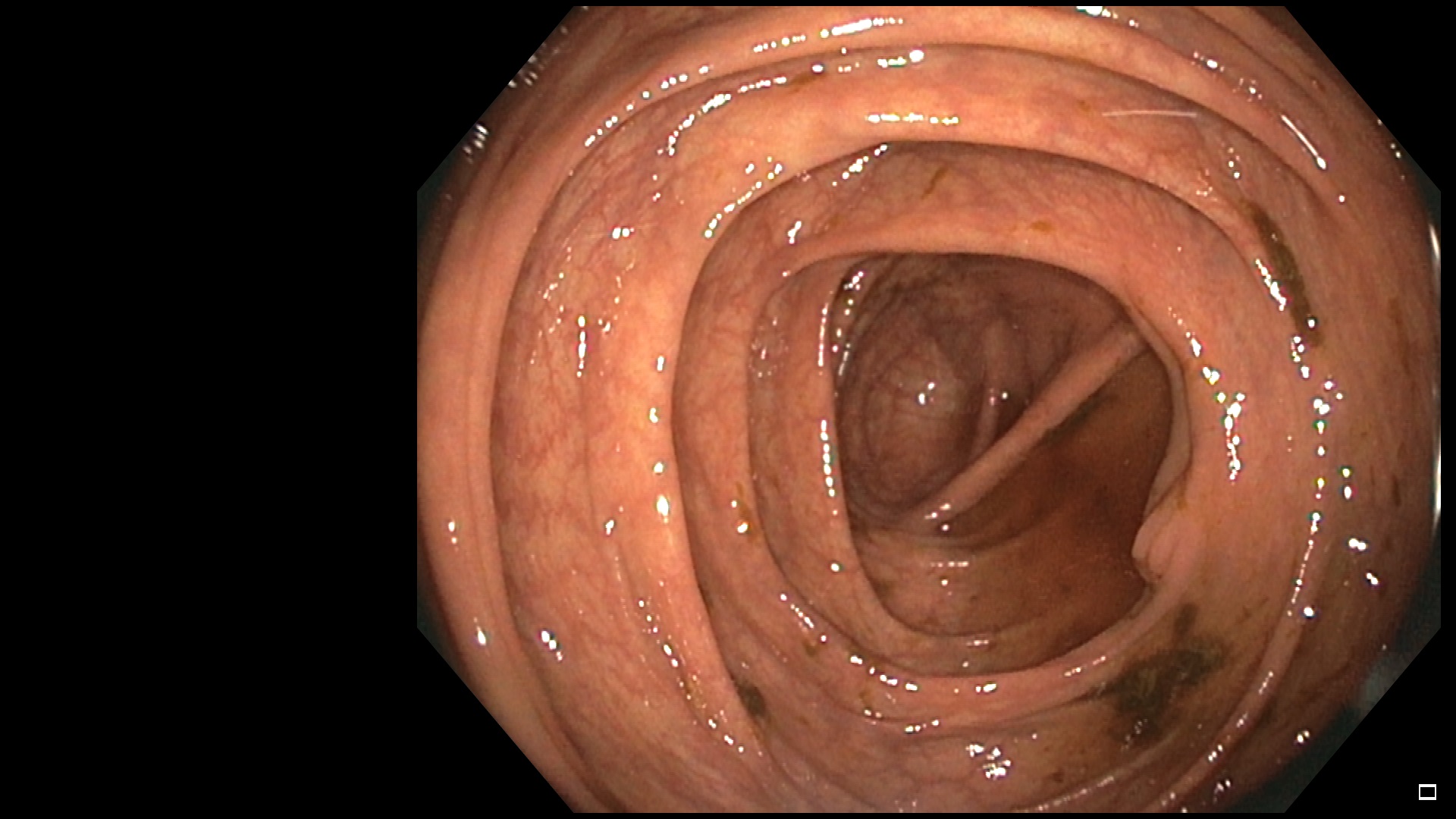}
  \caption{}
  \label{fig:sample_cek}
\end{subfigure}
\begin{subfigure}[b]{0.3\textwidth}
  \centering
  \includegraphics[width=.7\linewidth]{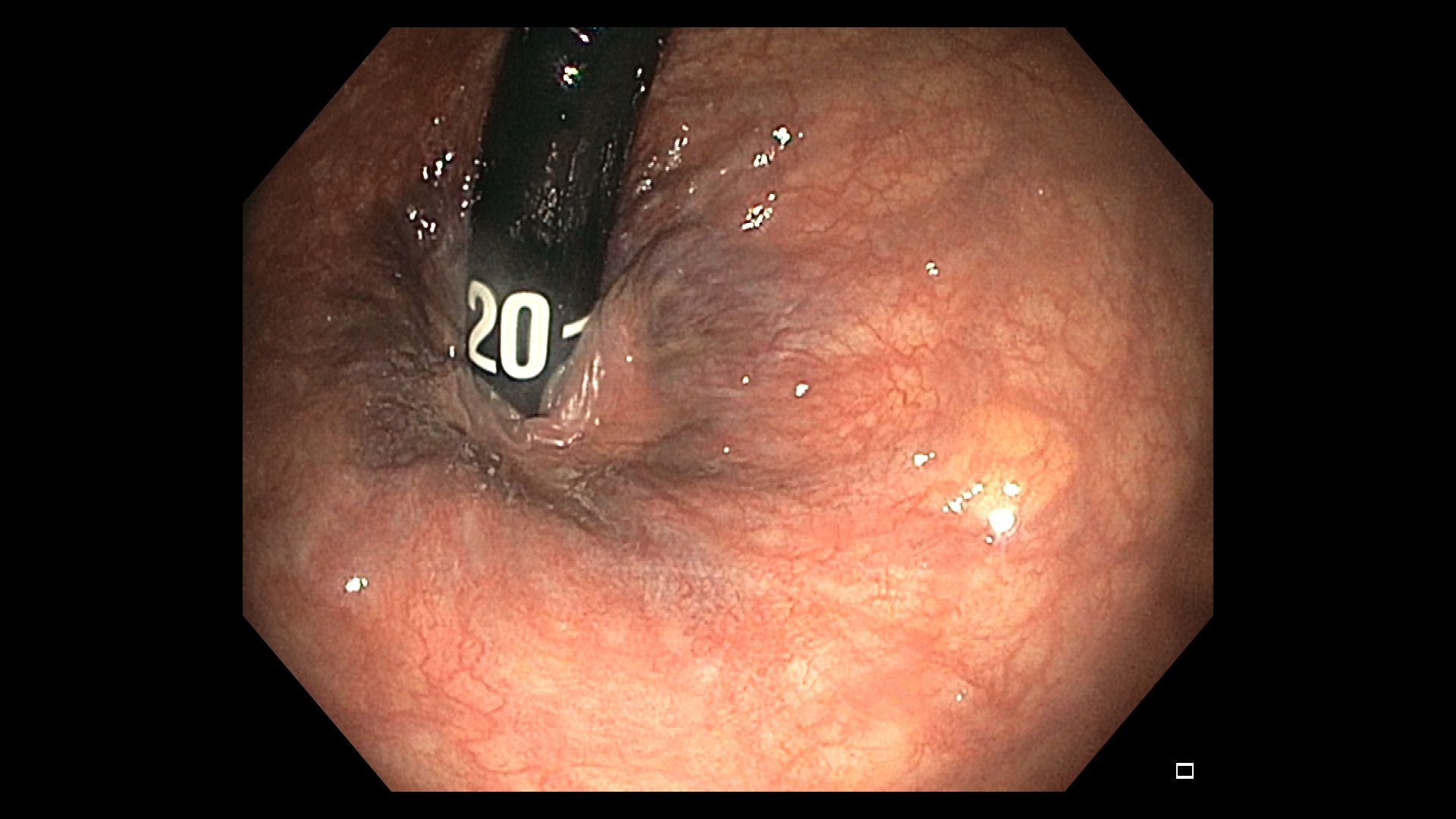}
  \caption{}
  \label{fig:sample_recrf}
\end{subfigure}
\begin{subfigure}[b]{0.3\textwidth}
  \centering
  \includegraphics[width=.7\linewidth]{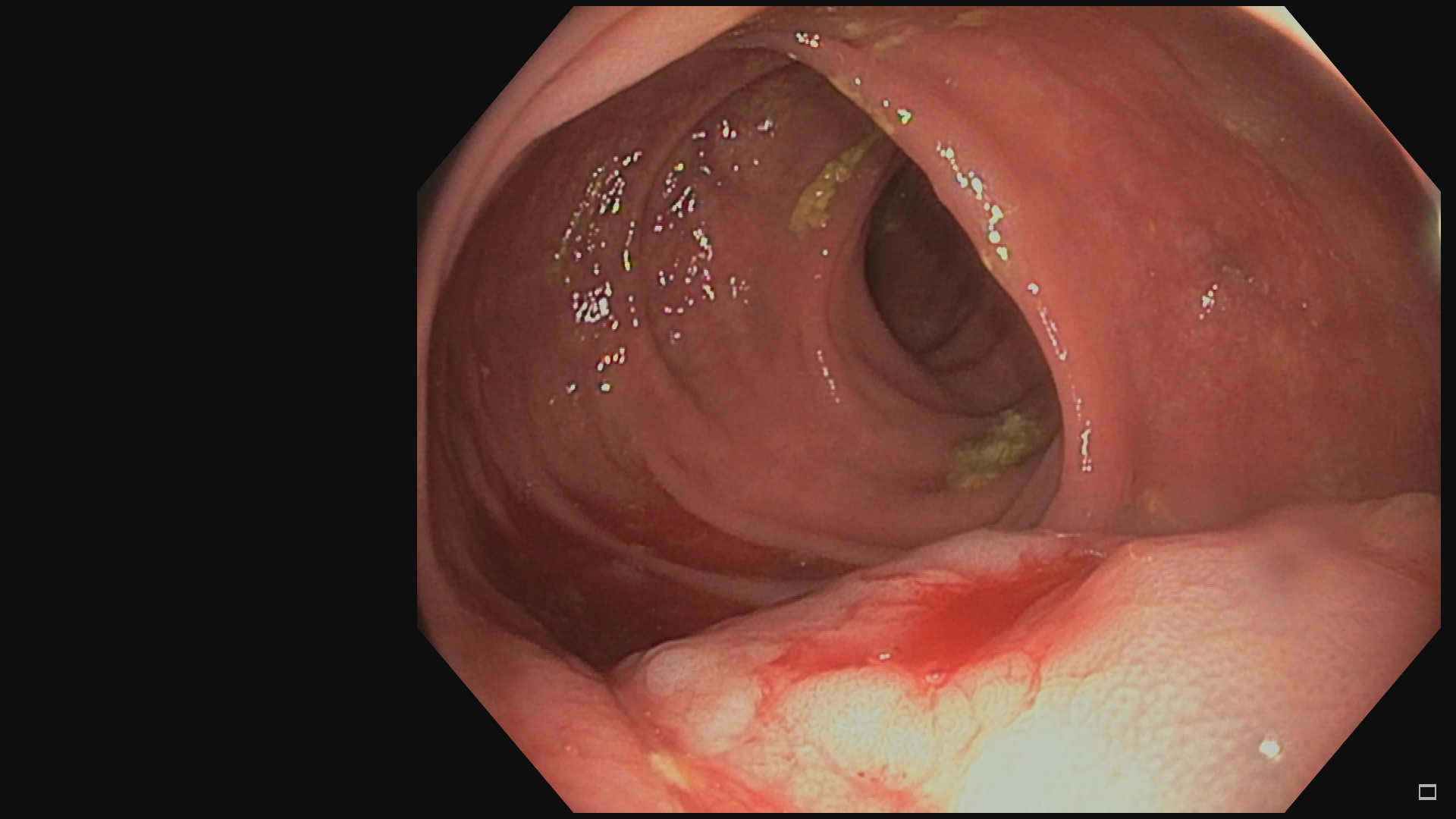}
  \caption{}
  \label{fig:sample_other}
\end{subfigure}
\caption{Sample snapshots with following annotations: Appendiceal Orifice (\ref{fig:sample_ao}), Ileocecal Valve (\ref{fig:sample_icv}), Cecum Landmark (\ref{fig:sample_cek}), Rectum Retroflexion (\ref{fig:sample_recrf}), Other (\ref{fig:sample_other})}
\label{fig:sample_snapshots}
\end{figure}

Our snapshots dataset contains 2416 snapshots collected from over 500 videos (separate from the training pool of 120 videos), identified and annotated by clinicians as described in Subsection \ref{sec:annot_cv}. A snapshot is a video frame that contains the anatomical/procedural feature of interest in reasonable focus, as identified by a medical specialist in a clinical setting. Each of the snapshots have been annotated according to the following labels: Appendiceal Orifice (AO), Ileocecal Valve (ICV)/ Cecum Landmark (Cec), Rectum Retroflexion (RecRF) and Other, which are shown with examples in Figure \ref{fig:sample_snapshots}. Since the Ileocecal Valve and the Cecum Landmark typically co-occur in snapshots due to their anatomical proximity, we combine them into a single label. Both of the first two labels describes the corresponding anatomical landmark. RecRF refers to the procedural action of retroflexion in the rectum i.e. bending the colonoscope backwards to inspect the rectum. Any other anatomical finding such as polyps, inflammation or general anatomical markers have been labelled as "Other". A breakdown of the number of frames per class has been given in Table \ref{snapshot_table}.

\subsection{Training dataset}
Our training dataset has 120 videos constituting of 2,000,000 frames in all, that were annotated and cross-validated as described in Subsection \ref{sec:annot_cv}. We face an extreme label imbalance, with a majority of frames ($>$ 95\%) belonging to a non-landmark (Other) class, and the minority containing a landmark of interest. We balance the dataset as part of our training and evaluation (to get a distribution similar to the snapshots dataset) as described in Section \ref{sec:train}.
\section{Problem definition}
We define our problem as follows: identify a function $f: C \times H \times W \xrightarrow{} J $ to  classify an image frame $F$ as one of the landmark classes $j \in \{\textrm{AO}, \textrm{ICV/Cec},$ $\textrm{RecRF}, \textrm{Other} \}$ such that $f(F_{ij}) = j, \quad \forall i \in \mathcal{S},  j \in J$. Here, $\mathcal{S}, J$ denote the set of snapshots and class labels respectively. We approximate $f$ using a deep neural network due to their proven capacity for modeling image data. We thus reduce our problem to finding the optimal weights $\theta^*$ for the following empirical loss $(\mathcal{L})$ :
\begin{equation}
    \theta^* = \arg \min_{\theta} \sum\limits_{i,j} \mathcal{L}(f(h(F_{ij})|\theta), j)
\end{equation}
Here, $h$ refers to a general data preprocessing function. Our framework supports any loss function $\mathcal{L}$ that is a distance metric between the predicted probability distribution and the true labels. Based on our experiments, we choose a Kullback-Leibler Divergence \cite{kim2021comparing} as the loss function $\mathcal{L}$.

\begin{figure}
\centering
\includegraphics[scale=0.3]{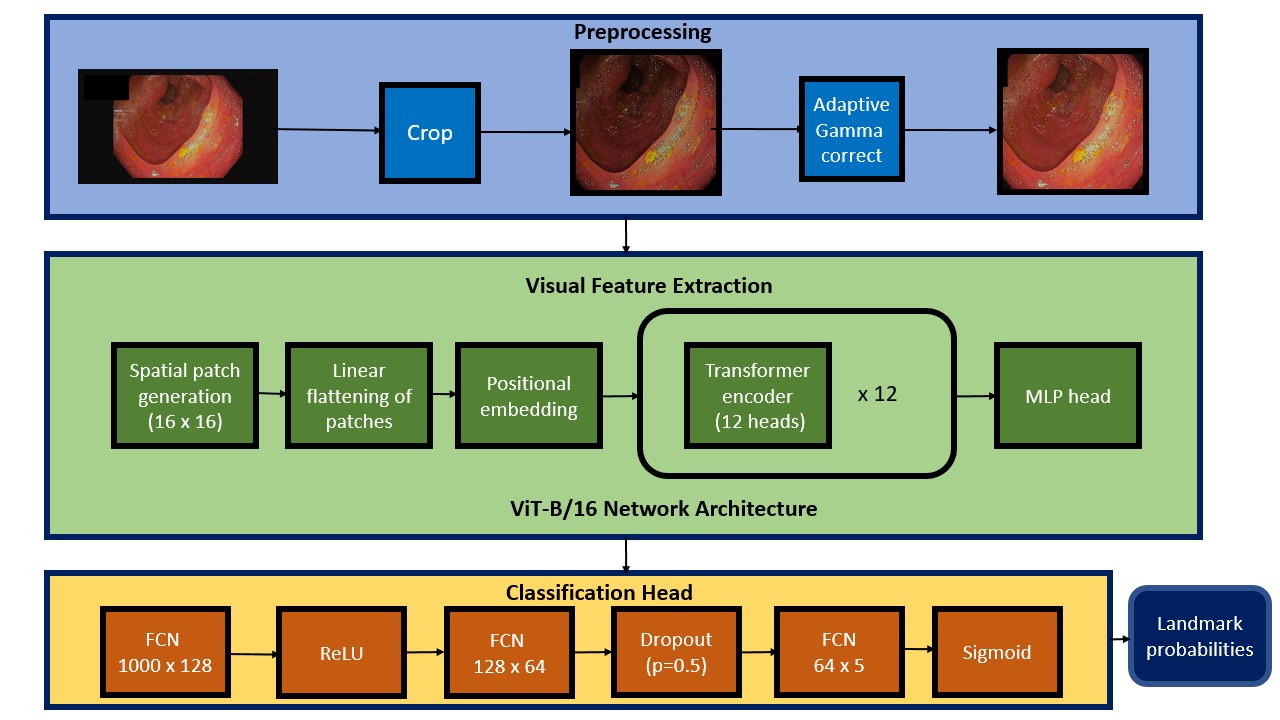}
\caption{Landmark detection pipeline architecture}
\label{fig:network_arch}
\end{figure}

\section{Architecture}
\label{sec:arch}
Our algorithm consists of three primary parts: \textit{1)} image preprocessing, \textit{2)} visual feature extraction and \textit{3)} classification head. The image preprocessing consists of an auto-cropping step to remove dark edges that are an artifact of the colonoscopy software itself, and auto-correct the brightness using gamma correction. Since the brightness varies considerably during a colonoscopy, we use an adaptive gamma correction algorithm described in \cite{rahman2016adaptive}.
We use a pretrained Vision Transformer (ViT-B/16) as the visual feature extraction backbone in our primary model. We also experiment with other CNN based architectures (ResNet101 and ConvNext-B) that were identically pretrained on the ImageNet dataset and benchmark their performances. Finally, we a use a Fully Connected Network (FCN) based classifier head to compute the label probabilities from the feature vector generated by the backbone. A high-level overview of the architecture is given in Figure \ref{fig:network_arch}.

\section{Training pipeline}
\label{sec:train}
We design our framework to train a model on annotated videos so that it performs well on clinically selected snapshots. Snapshots are different from video frames because they are hand-picked by clinicians in the following regards: they have a different distribution of landmarks and have a different photometric quality. We address this gap in the training and evaluation data using:
\begin{enumerate}
    \item \textit{Cross-validation}: Cross-validating frames as explained in Section \ref{sec:annot_cv} reduces the possibility of annotation error and inclusion of poor quality frames in the training. This bridges the gap in data quality between snapshots and videos.
    \item \textit{Domain-specific sampling}: We artificially construct a training set that has a label distribution similar to the snapshots dataset by randomly sub-sampling the frames using a Bernoulli process, described in Equations \ref{eq: bernoulli}, \ref{eq: probability}. Thus, a frame $F_{ij}$ is included in the training set if $Z_{ij} = 1$. Here, $\mathcal{S}$, $\mathcal{T}$ are the snapshots and training sets respectively. $|\Gamma|$ denotes the cardinality of any set $\Gamma$.

\begin{equation}
\label{eq: bernoulli}
Z_{ij} \sim \textrm {Bernoulli}(p_j)
\end{equation}

\begin{equation}
\label{eq: probability}
p_j = min\left(\frac{\mid \bigcup_{i\in\mathcal{S}, k=j}F_{ik}\mid}{\mid \bigcup_{i\in\mathcal{S}, k}F_{ik} \mid} \middle / \frac{\mid \bigcup_{i\in\mathcal{T}, k=j}F_{ik}\mid}{\mid \bigcup_{i\in\mathcal{T}, k}F_{ik} \mid}, 1 \right)
\end{equation}
We repeat the sampling (with replacement) at the beginning of every epoch to maximally cover the downsampled frames.

    \item \textit{Sharpness-Aware Minimization Optimizer}: Learning anatomically relevant features and ignoring features generated by varying photometric conditions, specific clinical conditions etc. is critical to generalizability across multiple patient anatomies. We observe that using a SAM optimization scheme as described in \cite{foret2020sharpness} for training the neural networks helps learn such a robust model. 

\end{enumerate}

\section{Results}
\label{sec:res}
\begin{figure}
\centering
\begin{subfigure}[b]{0.3\textwidth}
  \centering
  \includegraphics[width=\linewidth]{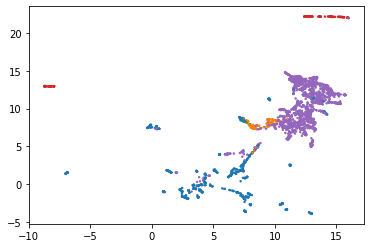}
  \caption{ViT-B/16}
  \label{fig:vit_umap}
\end{subfigure}
\begin{subfigure}[b]{0.3\textwidth}
  \centering
  \includegraphics[width=\linewidth]{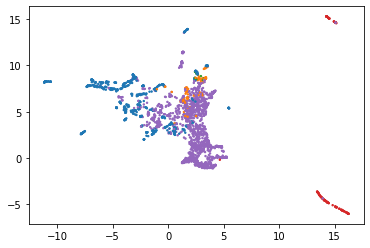}
  \caption{ConvNext-B}
  \label{fig:convnext_umap}
\end{subfigure}
\begin{subfigure}[b]{0.3\textwidth}
  \centering
  \includegraphics[width=\linewidth]{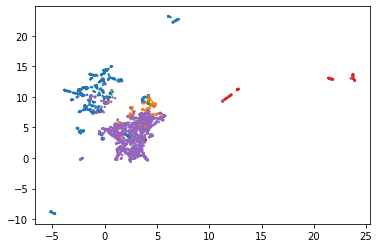}
  \caption{ResNet-101}
  \label{fig:resnet_umap}
\end{subfigure}
\caption{U-MAP embeddings of vision backbone representations with the color scheme: AO (Blue), ICV (Orange), Cec (Green), RecRF (Red), Purple (Other)}
\label{fig:umap}
\end{figure}

We have trained Vision Transformer (ViT-B/16), ResNet-101 and ConvNext-B based models using our framework and evaluated the results on our snapshots dataset. We tabulate the corresponding accuracy and the class-wise precision, recall scores in Table \ref{results_table}. We also plot 2D U-MAP \cite{mcinnes2018umap} embeddings of the vision backbone representations for images from our balanced test dataset in Figure \ref{fig:umap}. We report the test dataset statistics in Table \ref{snapshot_table}. We see that the vision transformer based model outperforms the other two on most metrics reported in Table \ref{results_table}. This is also corroborated by the comparatively well-separated clusters in Figure \ref{fig:umap}. We believe that the inherent shape bias of vision transformers, as reported in \cite{morrison2021exploring}, makes it more suitable than CNN-based architectures for landmark detection, since landmarks are reliably identified by their shape regardless of texture.

\begin{table}
\centering
\caption{Recall, precision scores and overall accuracy on snapshots dataset}\label{results_table}
\begin{tabular}{|l|l|l|l|l|}
\hline
Class & Metric & ViT-B/16 (Main) &  ResNet-101 & ConvNext-B\\
\hline
Overall & Accuracy & \textbf{81.84\%} & 73.06\% & 60.45\%\\
\hline
\multirow{2}{*}{AO} & recall &  68.15\% & 69.69\% & \textbf{75.09\%}\\
\cline{2-5}
& precision &  \textbf{76.41\%} & 55.36\% & 57.12\%\\
\hline
\multirow{2}{*}{ICV/Cec} & recall &  \textbf{89.43\%} & 75.33\% & 88.11\%\\
\cline{2-5}
 & precision &  51.26\% & \textbf{55.52\%} & 24.84\%\\
 \hline
\multirow{2}{*}{RecRF} & recall &  \textbf{96.09\%} & 86.31\% & 88.12\% \\
\cline{2-5}
 & precision &  \textbf{98.29\%}& 97.48\% & 95.03\%\\
 \hline
\multirow{2}{*}{Other} & recall &  \textbf{77.24\%} & 65.05\% & 28.39\%\\
\cline{2-5}
 & precision &  \textbf{85.10\%} & 74.48\% & 82.55\%\\
\hline
\end{tabular}
\end{table}

\section{Inference and Future Work}
\label{sec:inf}
We achieve an overall landmark classification accuracy of 81.84\% on a snapshot dataset of clinically relevant colon landmarks using a vision transformer backbone. We observe that a transformer based backbone outperforms other state-of-the-art CNN-based backbones such as ResNet-101 and ConvNext-B. We can visually see that well-separated representations on an independent, balanced test set imply a higher accuracy in Figure \ref{fig:umap}. This may be due to the transformer's inherently higher shape bias as reported by \cite{morrison2021exploring}. We hypothesize thus, since the accuracy trend is not completely explained by the number of parameters, with ViT-B/16 (86.6M) and ConvNext-B (89M) having a comparable number of parameters. 
\\ Further, the Rectum Retroflexion class has the highest precision and recall scores as well as the best separation on the U-MAP plots. This is because most RecRF frames are characterized by the presence of a metallic/plastic tube indicating the inversion of the colonoscope head. We further observe that the precision for AO and ICV classes is relatively lower. This is also evidenced by the poorer separation of the corresponding clusters in Figure \ref{fig:umap}. This can be explained by the visual similarity between these two landmarks and other parts of the colon (labelled "Other"), making it a challenging task. Thus, we can conclude from our results that detecting subtle anatomical features (such as a cecum landmark) as opposed to specific shapes (such as the retroflexion tube) is challenging for the vision backbone. 
\\ Finally, we believe incorporating temporal information in our future work will help remove false positives for both these classes and improve precision. We also believe that more complex training techniques such as active learning, self-supervised pre-training can further improve the quality of features learned by the vision backbone and improve accuracy. So, we plan on incorporating them in our future pipeline. We also plan on including more landmark classes such as polyps and diverticula in the future.

%
%
%
%
%
\bibliographystyle{splncs04}
\bibliography{literature_review}
%





\end{document}